# Human Activity Recognition Using Visual Object Detection

Schalk Wilhelm Pienaar[1], Reza Malekian[1,2], Senior Member, IEEE,
[1]Department of Electrical, Electronic and Computer Engineering, University of Pretoria, Pretoria,0002, South Africa

[2]Department of Computer Science and Media Technology, Malmö University, Malmö, 20506, Sweden/ Internet of Things and People Research Center, Malmö University, Malmö, 20506, Sweden
Reza.malekian@ieee.org

*Abstract*—Visual Human Activity Recognition (HAR) and data fusion with other sensors can help us at tracking the behavior and activity of underground miners with little obstruction. Existing models, such as Single Shot Detector (SSD), trained on the Common Objects in Context (COCO) dataset is used in this paper to detect the current state of a miner, such as an injured miner vs a non-injured miner. Tensorflow is used for the abstraction layer of implementing machine learning algorithms, and although it uses Python to deal with nodes and tensors, the actual algorithms run on C++ libraries, providing a good balance between performance and speed of development. The paper further discusses evaluation methods for determining the accuracy of the machine-learning and an approach to increase the accuracy of the detected activity/state of people in a mining environment, by means of data fusion.

Keywords— Activity Recognition, Acceleration Sensors, Augmentation, Common Objects in Context, Data Fusion, Object Detection, Tensorflow

## I. INTRODUCTION

With the development of multiple types of sensors and techniques used for monitoring the safety of underground miners, the scope and number of different types of sensor data have become more available to perform scientific research on. With this increasing growth, the need to combine data from these different sources has become increasingly in demand, in order to have the most useful information extracted. Data fusion techniques are found to be an effective way to utilize the information from such large sets of data. It paves the way for enhancing a system's understanding of the surrounding environment and the ability to automate a decision-making system. Data fusion can be applied in the fields such as visual enhancements, object detection, and area surveillance [1].

Neural networks have the ability to solve machine-learning problems with large datasets and are proven to be excellent at learning from training datasets and creating models from this data. Tensorflow is an open source framework that has the ability to create such models from a given neural network and can be incorporated to do object detection from a real-time video stream. Tensorflow has also models that can be used as the base for transfer-learning, for better accuracy, considering a much shorter time. Once the system is trained to properly detect different entities (i.e. people and objects) within a mining environment, it can be further trained to track its activity or get the context of what is happening within a mining environment at any given time. Together with other sensors, the implementation of data-fusion would further assist with the contextual information of the surrounding environment.  For instance, if a camera records the event of an underground miner lying down, versus falling, the data can be fused with an accelerometer attached to the miner to determine if it was an actual fall or just lying down, without complex methods that do so based on visual data only. If the camera loses sight to the miners (due to dust, fire, collapsed mine, or simply something insignificant covering the camera's line of sight) it may be necessary to combine this knowledge with a sound pressure level meter, temperature sensor, gas sensor and possibly other environmental sensors. The activity of miners (heart-rates, movements, etc.), obtained from wearable health monitoring systems, can also be part of the analysis.

## II. DATA FUSION METHODS

Data fusion becomes necessary in situations where a system is not fully observable (i.e. the system's current state cannot be inferred from individual output sensors). Depending on the context, it may even become necessary to fuse sensor data with external sensors, such as a wearable HAR device together with an external observing camera.

For machine learning classifiers, the algorithms work on a vector of given attributes that define an object or process. With data fusion, where multiple domains of data need to be combined, not all data is always present at all times, and the data obtained from the various sensors are not always in the correct format to be used directly, so there is a need for pre-processing. In order to have the data combined, it is therefore required to convert the various data points into a common data set. This is known as feature fusion data alignment [2]. The process involves finding synonyms, or otherwise known as semantic processing [3], which is basically finding data that have the same meaning, but from different sensor types. The second required process involves domain conversions, which reduces the number of different attribute values into a discrete set of values, which is typically found in classifiers.

Various methods for the fusion of visual data and data collected from and Inertial measurement unit (IMU) are discussed in [4].

## III. MODEL TRAINING METHODS

In order to train our model that is able to detect, for instance, an injured miner, it is necessary to follow a couple of steps, namely:

### A. Collection of images

Collection of images that involves the scenario or entity of interest. A couple of hundred such images would be ideal, but the more images, the more accurate results are expected. Typically, about 70-90% will be used for training a model, and about 10-30% for testing. The training and testing data should consider pictures/frames from completely different environments to ensure independence. In order to ensure variability, background environments and angles of pictures should change. For instance, photos need to be taken from all sides of the entity of interest. The data for our purpose will consist of:

| Image category | Quantity |
|---|---|
| Testing | 220 |
| Training | 19 |
| Evaluation | 19 |

Each picture has a possible varying number of labelled entities.

*A. Image Augmentations*

For further enhancement of the variability of each image, image augmentation needs to be performed. This includes translating, rotating, scaling, flipping etc. of images. Convolutional neural networks that have the ability to accurately classify objects, placed differently, are considered to be invariant [5]. Tensorflow provides the ability to configure the augmentation, with the following configurations that are available, and the ability to set values such as min and max:

```
NormalizeImage
RandomHorizontalFlip
RandomPixelValueScale
RandomImageScale
RandomRGBtoGray
RandomAdjustBrightness
RandomAdjustContrast
RandomAdjustHue
RandomAdjustSaturation
RandomDistortColor
RandomJitterBoxes
RandomCropImage
RandomPadImage
RandomCropPadImage
RandomCropToAspectRatio
RandomBlackPatches
RandomResizeMethod
ScaleBoxesToPixelCoordinates
ResizeImage
SubtractChannelMean
SSDRandomCrop
SSDRandomCropPad
SSDRandomCropFixedAspectRatio
SSDRandomCropPadFixedAspectRatio
RandomVerticalFlip
RandomRotation90
RGBtoGray
ConvertClassLogitsToSoftmax
RandomAbsolutePadImage
RandomSelfConcatImage
```

A full up-to-date list can be found in Tensorflow's *preprocessor.proto* file.

*B. Image Labelling*

The next step involved is the categorization of our data. I.e., each of our images should be labelled and a box drawn around the area of interest. For Tensorflow, this requires a Pascal VOC XML file that defines the area of interest in each given image.

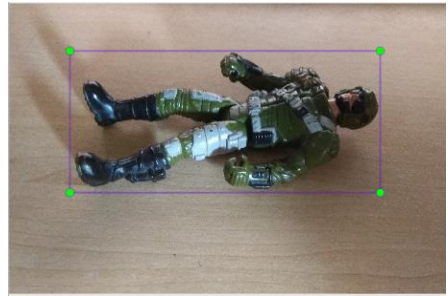

Figure 1: labelling images for training

The resulting Pascal VOC format file contains information such as the difficulty (e.g. is the item obstructed), the image dimensions and file name, the bounding box dimensions and the label.

From the labelled images, two CSV files are to be set-up; one for test data and another for training. The following information is extracted for each image, from the generated XML file. This information is then used for the step where TF-generating step.

Table 1: Required data for setting up TF-records

| file | w | h | class | x-min | y-min | x-max | y-max |
|---|---|---|---|---|---|---|---|
| abc.jpg | 1067 | 1600 | lying | 1 | 504 | 989 | 1240 |
| xyz.jpg | 1080 | 1080 | standing | 21 | 184 | 1062 | 1066 |

*C. Create a Label map file*

A label map file needs to be set-up; where all the labelled entities with their names, id's and display names are added, in a JSON-like format. The names defined should match what is defined in the VOC files, so that the TF file-generating script can match the names and link it with an ID.

*D. Generate the TF file*

Test and training data are separated and TF-records can be generated from these files. A TF file is a binary file format that is able to combine multiple datasets and provides a performance benefit compared to individual images, since it is more light-weight, thus using less disk space and can be read faster. The resulting files would be in a similar structure as shown below, allowing us to generate the TF file. The script used to generate these records are provided by the object detection dataset tools. The one that is applicable for the COCO dataset is found in [6]

```
-data/
--test_labels.csv
--test.record
--train_labels.csv
--train.record
--object-detection.pbtxt
-images/
--test/
---[image<x>.jpg]
--train/
---[image<x>.jpg]
-training
--ssd_mobilenet_v1.config
```

*E. Learning/Transfer learn model*

Transfer learning from existing models is beneficial as it could take weeks to train a model from scratch, even with high-end hardware. Transferring allows us to re-use data from proved learning. The process borrows pre-trained parameters, and therefore makes it possible to reach our target accuracy much quicker [7]. An SSD model is used with MobileNet and the COCO dataset (ssd_mobilenet_coco). The SSD Mobilenet configuration file will be used as the basis for continuing with the model training. If necessary, additional image augmentation parameters will be added, which is not already part of the default configuration. It should be noted that the existing model has a mean average precision (mAP) of 21 on the data set it was trained on (COCO) [8].

MobileNet is a base network that provides high-level features for classification/detection; the architecture is more suitable for mobile and embedded vision applications as there is some sacrifice on accuracy, for performance [9]. SSD is a single shot detector, which means that when detecting, only a single image is required to detect multiple objects. The COCO dataset, that can be used to transfer-learn this model, consists of 328 000 images, of which there are 2.5 million labelled instances, and about 90 categories [10].

In order to transfer-learn an existing model, it is necessary to continue from the trained models' last checkpoint. These checkpoints are generated every few hundred steps in the training process, and once the results are satisfactory, an inference graph can be used for object detection. Since new data/knowledge is being added in terms of a new category, the layer which needs to be retrained should be specified, otherwise all the layers and their weightings will be adjusted. The script used to train and evaluate the training model is provided together with Tensorflow's research tools [11].

*F. Training Analysis*

It is important to analyze the accuracy of the trained model. There are various such methods that can be used for the analyses, including IoU, mAP, precision, recall, etc., and are discussed below. To make sense of the below, it is necessary to understand the meanings of true-positive (TP), false-positive (FP), and false-negative (FN) in terms of Tensorflow;

**TP:** A true-positive is defined as a detection where the IoU is greater than 50%, and in the case of multiple boxes being detected for a single entity, a TP would be the box with a greater confidence level.

**FP:** A false positive is the predicted boxes that have been preceded by those with greater confidence. Or When an entity is completely incorrectly labelled and it's the only one detected.

**FN:** An FN is an entity that is not detected at all, even though it has been trained for [12].

*G. Intersection over union (IoU)*

IoU is an evaluation metric that can be applied to any predicted bounding boxes where the ground-truth (the manually labelled bounding boxes) is known. IoU is calculated by dividing the overlapping area between the bounding boxes by the area of the union. I.e, the area that is common between the two bounding boxes over the area occupied by the predicted box and ground-truth box [13]:

$$IoU = \frac{Area\ of\ overlap}{Area\ of\ union} \quad (1)$$

An intersection score that is above 0.5 is considered good for a prediction.

The importance of IoU as a metric for determining whether a prediction is good, is because it does not work in a binary fashion, but rather as a level of accuracy; it is unlikely to find an exact match between the prediction and ground-truth boxes. An IoU of close to 1 means that the prediction is close to perfect. The meaning of the graph in Figure 2, below, which shows the IoU graph of our training model as a function of IoU over the number of steps, is the mean average precision of the models' correct prediction percentage for all the trained classes. From the graph, it can clearly be seen that as the number of steps increases, the IoU start reaching a value closer to 1.

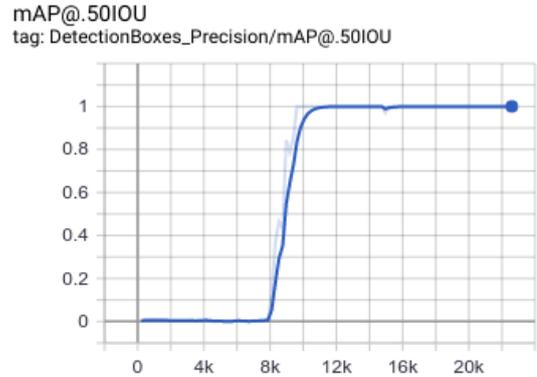

Figure 2: Development of mAP model training

*Mean average precision (mAP):*

mAP is a widely used metric for model performance comparisons in object detection. Precision-recall curves summarise mAP information into a single number. Such function is formed by sorting the predicted bounding boxes of the entire set of evaluation-images, by their class and confidence, then a recall and precision value is calculated for each prediction. The average precision is then calculated from the area under the curve [14]. mAP is the mean value of AP values but across all classes.

*Precision:*

Precision is a measurement of how relevant the detection results are, and is given by the following equation [12]:

$$precision = \frac{TP}{TP + FP} \quad (2)$$

*Recall:*

The percentage of relevant objects detected are described by the recall function. The formula to calculate this is defined as [13]:

$$recall = \frac{TP}{TP + FN} \quad (3)$$

Figure 3, below, shows the average recall graph for a set of 100 objects in the dataset. A recall value of 1 indicates that all the objects in the analysed dataset were correctly predicted. Therefore, it is seen that the training process only started showing positive results after about 8000 steps. Since only large images were trained, this graph does not necessarily reflect the detection of medium to small images; the graphs for medium to small objects both got recall values of -1, which means that data is absent for those sizes.

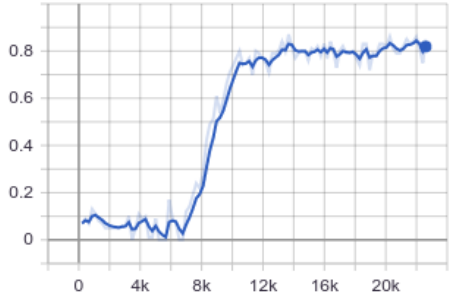
Figure 3: Average recall for 100 objects

*F1-score:*

It should be noted that an increase in precision generally means that the recall decreases, or the other-way around. I.e. with a low threshold value, lots of false-positives are detected, whereas with a high threshold value, not as much false-positives are detected, but typically results in certain entities not being detected at all. The metric for combining these two attributes is known as the F1-score. The equation is given below [15]:

$$F1_{score} = 2 \times \frac{precision \times recall}{precision + recall} \quad (4)$$

Figure 4 below shows the training result of the classification loss function versus computational steps. This is a computationally feasible loss function that indicates the price paid for the inaccuracy of the predictions made in classification-problems [16]. The lower the loss value, the better trained the model is. The graph gives an overview of how the training algorithm attempts to reduce the losses, up to a point where the losses are satisfactorily low.

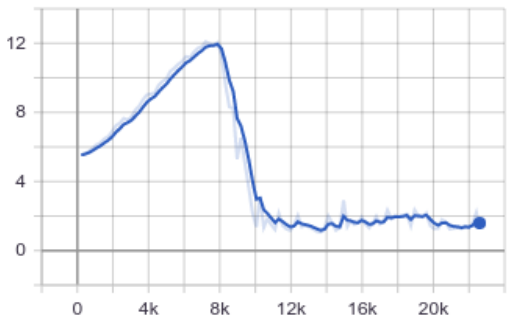
Figure 4: Classification Loss vs steps

Figure 5 shows the training result of the localization loss function. The same as the classification loss, this is a computationally feasible loss function that indicates the price paid for the inaccuracy of the predictions, but for localization-problems instead of classification-problems. This graph also indicates loss vs steps.

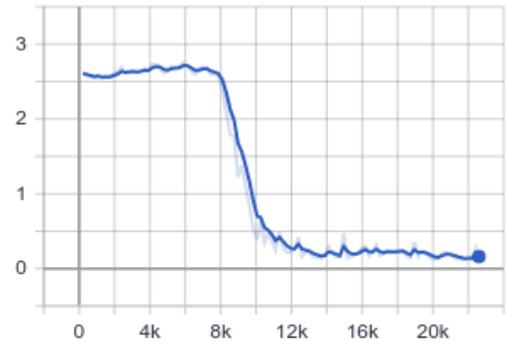
Figure 5: Localization loss vs steps

Figure 6 below shows the training result of the total loss function. This is an indication of the total number of losses, and the goal would typically be for the training algorithms to reach a loss of about 1%. It may, however, end up that a few hundred thousand more steps do not necessarily result in a lower loss-percentage.

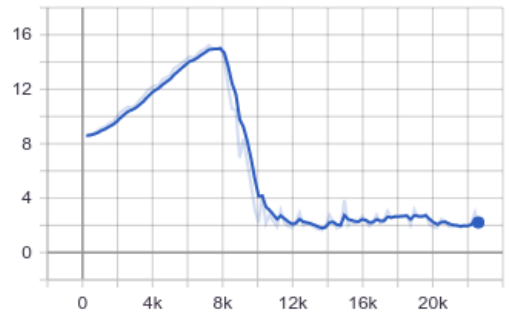
Figure 6: Total loss vs steps

*H. Object Detection*

After having satisfactory training results, it is possible to export a Tensorflow graph proto file from a certain checkpoint, which can finally be used for object detection. Each checkpoint reflects different losses, etc. The script used to evaluate the object detection can be found in [17].

IV. RESULTS

Figure 7 shows the result of the trained object detector [18,19], predicting [20,21] a standing toy soldier, on an untrained image.

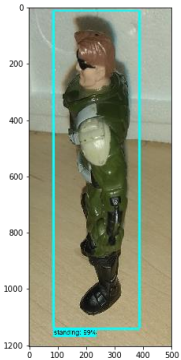
Figure 7: Detection un-trained toy-soldier standing, 89% confidence

Figure 8 shows the result of the trained object detector, predicting a person lying down, on an untrained image [22].

The training images set also did not include an image of real human beings [23], yet has the ability to detect one .

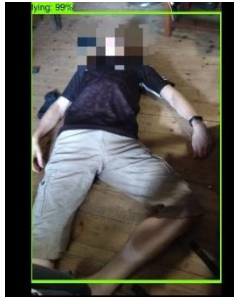

Figure 8: Detection of untrained human, lying-down, with 99% confidence

Figure 9 shows the correction prediction of a toy-soldier, lying face to the ground.

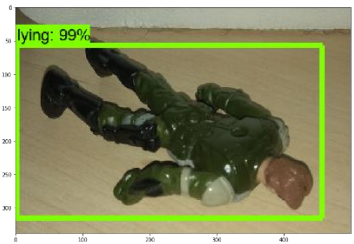

Figure 9: Detection un-trained toy-soldier lying-down, 99% confidence

## V. Conclusion

Model training was analysed by both fine-tuning and learning models from scratch. Fine tuning allowed objects to be trained relatively quickly (a few hours on a CPU), and detected objects with great confidence. The problem, however, is that it would do lots of false-positive detections; i.e. random other types of objects were detected as either standing-up or lying-down (i.e. the two classes that have been trained), along with the actual object. In an attempt to solve this, training images were presented with completely different backgrounds and image augmentation was implemented in order to learn the model to distinguish an object from different backgrounds (i.e. negative training). This showed better results, but some difficulty was still experienced with a couple of false-positive detections. Training the model from scratch with various backgrounds was then attempted - this showed even better results, although there was still some difficulty with lots of false-positive detections on a human shoulder, which added a couple of "standing" or "lying" down bounding boxes. For other types of objects, little false-positives were detected. This is, however, something that should be solved with more training steps.

Training models using other methods, such as SSD Inception, and RCNN proved to be more accurate, but since these algorithms have not been optimised for mobile devices, it was found to be too process-intensive to run on a CPU, especially with a live video feed. Training and detecting using an Nvidia GPU supporting CUDA would have made this an option.

The various graphs and training analyses methods that have been analysed and discussed in this paper could be used as a reference point for future model training research.